\documentclass{llncs}
\usepackage[utf8]{inputenc}
\usepackage{graphicx}
\usepackage{amsmath}
\usepackage{subcaption}
\title{Stack-U-Net: Refinement Network for Improved Optic Disc and Cup Image Segmentation}
\author{Artem Sevastopolsky\inst{1,2} \and Stepan Drapak\inst{1,3} \and Konstantin Kiselev\inst{1} \and Blake M. Snyder\inst{4,5} \and Jeremy D. Keenan\inst{5,6} \and Anastasia Georgievskaya\inst{1,7}}
\institute{Youth Laboratories Ltd., Moscow, Russia \and 
Skolkovo Institute of Science and Technology, Moscow, Russia \and
Lomonosov Moscow State University, Moscow, Russia \and
University of Colorado Denver School of Medicine, Aurora, CO, USA \and 
Francis I. Proctor Foundation, University of California San Francisco, San Francisco, CA, USA \and
Deparment of Ophthalmology, University of California San Francisco, San Francisco, CA, USA \and
Institution of Russian Academy of Sciences Dorodnicyn Computing Centre of RAS, Moscow, Russia}
\begin{document}
\maketitle
\begin{abstract}
In this work, we propose a special cascade network for image segmentation, which is based on the U-Net networks as building blocks and the idea of the iterative refinement. The model was mainly applied to achieve higher recognition quality for the task of finding borders of the optic disc and cup, which are relevant to the presence of glaucoma. Compared to a single U-Net and the state-of-the-art methods for the investigated tasks, the presented method outperforms others by multiple benchmarks without a need for increasing the volume of datasets. Our experiments include comparison with the best-known methods on publicly available databases DRIONS-DB, RIM-ONE v.3, DRISHTI-GS, and evaluation on a private data set collected in collaboration with University of California San Francisco Medical School. 
The analysis of the architecture details is presented. It is argued that the model can be employed for a broad scope of image segmentation problems of similar nature.

\end{abstract}

\section{Introduction}


Glaucoma is the second leading cause of blindness all over the world, with approximately 60 million cases reported worldwide in 2010, and an increase by 20 million is expected in 2020 \cite{med_almazroa,med_quigley}. If left unnoticed, glaucoma can cause irreversible damage to the optic nerve leading to blindness. Therefore, diagnosing glaucoma at early stages is very important \cite{med_almazroa}. 

Optic nerve examination includes eye fundus test, which requires a doctor localizing areas of optic disc and optic cup (central part of optic disc) and finding their borders. Presence of glaucoma can be identified by noticing optic nerve cupping, i.e. increase of optic cup in size. One of the main indicators of the disease is cup-to-disc ratio (CDR) --- a ratio between heights of cup and disc \cite{med_almazroa}. It is considered one of the most representative features of optic disc and cup areas for glaucoma detection, and, according to \cite{akram}, eye with CDR of at least 0.65 is usually considered as glaucomatous in clinical practice.

Relative size of these two organs is one of the most valuable factors determining the presence of glaucoma.  Segmentation of the optic disc and cup is a very time-consuming task currently performed only by the professionals. As stated in \cite{lim2015integrated}, according to the research, full segmentation of optic disc and cup requires about eight minutes per eye for a skilled grader. Solutions for automated analysis and assessment of glaucoma can be very valuable in various situations, such as mass screening and medical care in countries with significant lack of qualified experts.


Computer-aided diagnosis of glaucoma can be based on the optic disc and cup segmentation algorithms. Nowadays, methods of deep learning provide the state-of-the-art results on many tasks of image processing, including the semantic and instance segmentation. In many cases, a small number of objects is to be found, but, on the other hand, often only small datasets can be acquired, class imbalance is present, and very high recognition quality and robustness is required \cite{milletari2016v}.

In this work we intend to provide a new end-to-end approach to the medical segmentation task of optic disc and cup borders localization, which is based on well-known and highly-performing U-Net \cite{ronneberger2015u} convolutional neural network (CNN) of encoder-decoder style. The latter is used as a basic block for a cascade of networks employed as the main model proposed. We refer to the neural network built as Stack-U-Net. 

Compared to many other approaches of building the cascade of refinement networks, the one proposed in this work does not depend on the structure of the task and can be straightforwardly applied to many applications of image segmentation, image-to-image translation, etc.

Despite the linear growth of the number of parameters with the number of blocks, we observe that the model leads to the rate of overfitting similar to the original U-Net and only provides a noticeable quality gap. We consider this a consequence of regularly placed bottlenecks --- the first layers of each basic network. This way, the basic models, conditioned by an input image, are only working to refine the output of the preceding basic models. In this article we evaluate how the described extension can be employed to enhance image segmentation quality, and how many basic modules are optimal to make the full cascade learn hierarchy of representative features of an image. 

\section{Related work}


The idea of the cascade network is present in a large number of  various computer vision works. However, the information passed between sub-networks in a cascade is usually chosen differently and is sometimes implied by the structure of a solved problem.

The paper \cite{lin2017refinenet} applies a cascade multi-path refinement network by augmenting ResNet \cite{resnet} pretrained on ImageNet \cite{imagenet} with RefineNet blocks, which take the output of ResNet’s intermediate layers as an input and are organized in a decoder-like topology. Cascades of up to 4 2-scale RefineNet's are compared for the semantic segmentation problem. 
Similar approach is proposed in \cite{dai2016instance} for the task of instance-aware semantic segmentation: the first sub-network finds box instances (ROIs), they are fed to another sub-network which outputs a binary segmentation mask, and the mask is fed to another sub-network which segments separate instances.
	
In \cite{christ2016automatic}, two U-Net’s is applied for the liver and lesion segmentation in CT images as a model backbone, which is followed by 3D Conditional Random Field. Followed by the fact that the lesions are smaller regions inside the liver, the cascade is applied as follows: the first U-Net segments the liver, then its localized ROI is passed to a second U-Net. It is experimentally shown in the work that the Dice score can be improved this way by 20\% compared to a single U-Net. The same approach is applied in \cite{sevastopolsky2017optic} for the segmentation of the optic disc and the optic cup, as the latter is smaller than the optic disc and is always inside of it.

There is a number of works that apply cascade of neural networks in a fashion more similar to our proposed idea. For instance, in \cite{toshev2014deeppose} a well-known DeepPose method for human pose estimation is proposed, which is based on a cascade of regressors, iteratively refining each other. The first basic network localizes all the "skeleton" joints on an input image, and all the subsequent basic networks are refining previously found joints locations, conditioned by sub-images cropped by joints areas found. The work \cite{trigeorgis2016mnemonic} follows a close approach for the face landmarks detection, but also benefits an idea of applying recurrent neural network (RNN): the weights of all basic networks, starting from the second one, are shared, and the whole model is trained as the RNN.

\section{Stack-U-Net}

As a preprocessing, unsupervised Contrast-Limited Adaptive Histogram Equalization (CLAHE) \cite{szeliski2010computer} is applied in order to bring the brightness characteristics closer across all the dataset. 

\begin{figure}[ht!]
	\centering
    \includegraphics[width=1.0\linewidth]{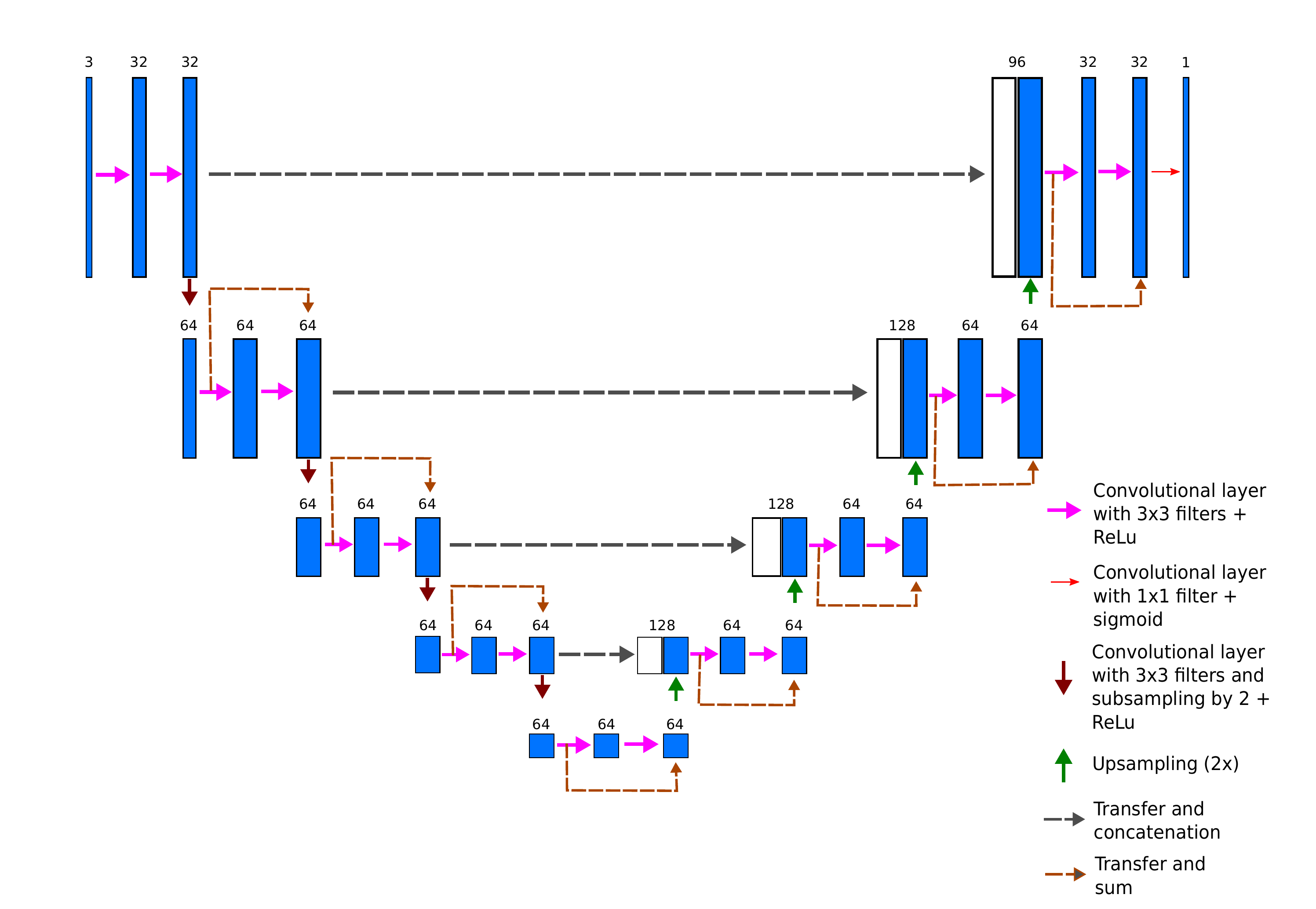}
    \caption{Res-U-Net architecture --- a basic block of the Stack-U-Net model. Another possible basic block is U-Net, which is the same module without residual connections marked light-brown dashed lines.}
    \label{fig:basic_block}
\end{figure}

\begin{figure}[ht!]
	\centering
    \includegraphics[width=1.0\linewidth]{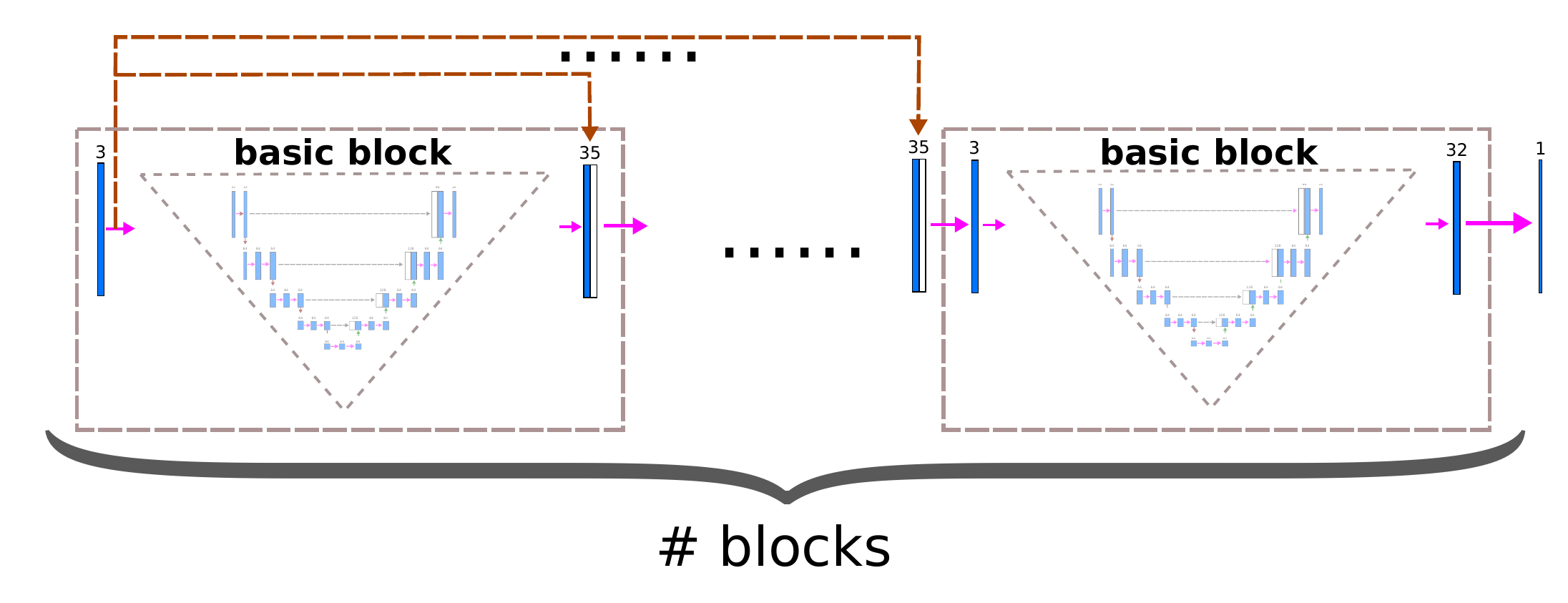}
    \caption{Stack-U-Net --- the main proposed model}
    \label{fig:stack-u-net}
\end{figure}

The presented cascade model, which we refer to as Stack-U-Net, is depicted on Fig.~\ref{fig:stack-u-net}. It consists of basic blocks, and each of them follows the encoder-decoder architecture similar to U-Net~\cite{ronneberger2015u}, depicted on Fig.~\ref{fig:basic_block}. We consider 2 kinds of basic blocks: U-Net and Res-U-Net. They both feature skip connections (shown gray on the Fig.~\ref{fig:basic_block}), linking layers of the encoder and decoder, which are of very high importance. Compared to the conventional U-Net, Res-U-Net also features residual connections (shown dashed light-brown on Fig.~\ref{fig:basic_block}). All the basic blocks except the last one, end with 32 feature maps, which are stacked with the input image by long skip connections (shown dashed light-brown on the Fig.~\ref{fig:stack-u-net}). The latter provide an additional information to the next basic block, so that it refines the previous features by directly accessing colors from the input image.  One can notice that Stack-U-Net with Res-U-Net blocks allows for relatively more efficient gradient propagation in terms of information, as it preserves an identity mapping \cite{he2016identity,lin2017refinenet} between input and output without any intermediate layers. 

  As a loss function, we use $l(A, B)$:
	\begin{gather*}
	l(A, B) = -\log d(A, B), \text{where:}\\
	d(A, B) = \frac{2\sum\limits_{i, j} a_{ij} b_{ij}}
	{\sum\limits_{i, j} a_{ij}^2 + \sum\limits_{i, j} b_{ij}^2},
	\end{gather*}
	where $A = (a_{ij})_{i=1\, j=1}^{H\;\;\;W}$ is a predicted output map, containing probabilities that each pixel belongs to the foreground, and $B = (b_{ij})_{i=1\, j=1}^{H\;\;\;W}$ is a correct binary output map. 
	
$d(A, B)$ is a real-valued extension of Dice score for binary images $Dice(A, B) = \frac{2|A \cap B|}{|A| + |B|}$. Along with Dice score, we report the Intersection-over-Union score values: $IOU(A, B) = \frac{|A \cap B|}{|A \cup B|}$, where $A$ and $B$ are defined as above. During the training, data augmentation was used to enlarge the training set by artificial examples. Images were subject to random rotations, zooms, shifts, flips and affine shears. Adam optimization method with learning rate of $10^{-5}$ was used.

\section{Experiments}

For experiments, we used the following datasets:
\begin{enumerate}
\item DRIONS-DB \cite{drionsdb} --- publicly available 110 color eye fundus images without cropping with annotation of the optic disc borders.
\item RIM-ONE v.3 \cite{rimone} --- publicly available 159 color eye fundus images with cropping (image side is approximately 5 times larger than the optic nerve diameter) with annotation of the optic disc and cup borders. Version 3 is the actual version.
\item DRISHTI-GS \cite{drishtigs1,drishtigs2} --- publicly available 50 color eye fundus images without cropping with annotation of the optic disc and cup borders. 
\item UCSF-DB --- private dataset of 963 color eye fundus images of 238 people without cropping, kindly provided by University of California, San Francisco (UCSF) Medical School, US and collected for optic disc and cup annotation tasks. For each photo, annotation of the optic disc and cup borders were prepared by 3 annotators. Final annotations were acquired as pixel-wise average of 3 masks for each of the 2 organs. Images were cropped by an~optic disc area (with gap of 20 pixels from each side) based on the ground truth annotations.
\end{enumerate}

For UCSF-DB dataset, several images of the same person were put either in train set altogether or in validation set altogether.

The comparison with the best found methods for the described public\linebreak databases is presented in Table~\ref{table:optic_disc_results} and Table~\ref{table:optic_cup_results}. We were unable to reproduce the results of other state-of-the-art methods. Evaluation on the large UCSF-DB dataset is presented in Table~\ref{table:ucsf_results}, which also contains a score of human annotator vs. another human annotator averaged by all pairs of annotators.

\begin{table}[ht!]
            \centering
        	\begin{tabular}{l|c|c|c|c|c|c}
        		\hline
        		& \multicolumn{2}{c|}{\,DRIONS-DB\,} &
        		\multicolumn{2}{c|}{\,RIM-ONE v.3\,} &
        		\multicolumn{2}{c}{\,DRISHTI-GS\,} \\
        		& IOU & Dice & IOU & Dice & IOU & Dice \\ \hline \hline
        		Stack-U-Net (15 ResU-Net blocks) & \textbf{0.92} & 0.96 & 0.91 & 0.95 & \textbf{0.95} & \textbf{0.97} \\
                Stack-U-Net (15 U-Net blocks) & 0.90 & 0.95 & \textbf{0.92} & \textbf{0.96} & 0.94 & \textbf{0.97} \\
                \hline
        		U-Net \cite{sevastopolsky2017optic} & 0.89 & 0.94 & 0.89 & 0.95 & 0.90 & 0.95 \\
                
        		Maninis et al. 2016 \cite{driu} & 0.88 & \textbf{0.97} & 0.89 & \textbf{0.96} & --- & --- \\
        		Zilly et al. 2017 \cite{zilly2017glaucoma} & --- & --- & 0.89 & 0.94 & 0.91 & \textbf{0.97} \\ \hline
        	\end{tabular}
            \newline\newline
        	\caption{Results for \textbf{optic disc} segmentation. "---" indicates that the result is not reported.}
            \label{table:optic_disc_results}
\end{table}

\begin{table}[ht!]
        	\centering
        	\begin{tabular}{l|c|c|c|c}
        		\hline
        		& \multicolumn{2}{c|}{\,DRISHTI-GS\,} &
        		\multicolumn{2}{c}{\,RIM-ONE v.3\,} \\   
        		& IOU & Dice & IOU & Dice \\ \hline \hline
        		Stack-U-Net (15 ResU-Net blocks) & 0.80 & \textbf{0.89} & 0.73 & \textbf{0.84} \\
                Stack-U-Net (15 U-Net blocks) & 0.77 & 0.86 & 0.72 & 0.83 \\  
                \hline
        		U-Net with cropping by OD region 
                \cite{sevastopolsky2017optic} & 0.75 & 0.85 & 0.69 & 0.82 \\
        		Zilly et al. 2017 \cite{zilly2017glaucoma} & 0.85 & 0.87 & \textbf{0.80} & 0.82 \\
        		Zilly et al. 2015 \cite{zilly2015boosting} & \textbf{0.86} & 0.83 & --- & --- \\ \hline
        	\end{tabular}
            \newline\newline
        	\caption{Results for \textbf{optic cup} segmentation. "---" indicates that the result is not reported.}
            \label{table:optic_cup_results}
\end{table}

      \begin{figure}[ht!]
        \centering
        \includegraphics[width=0.7\linewidth]{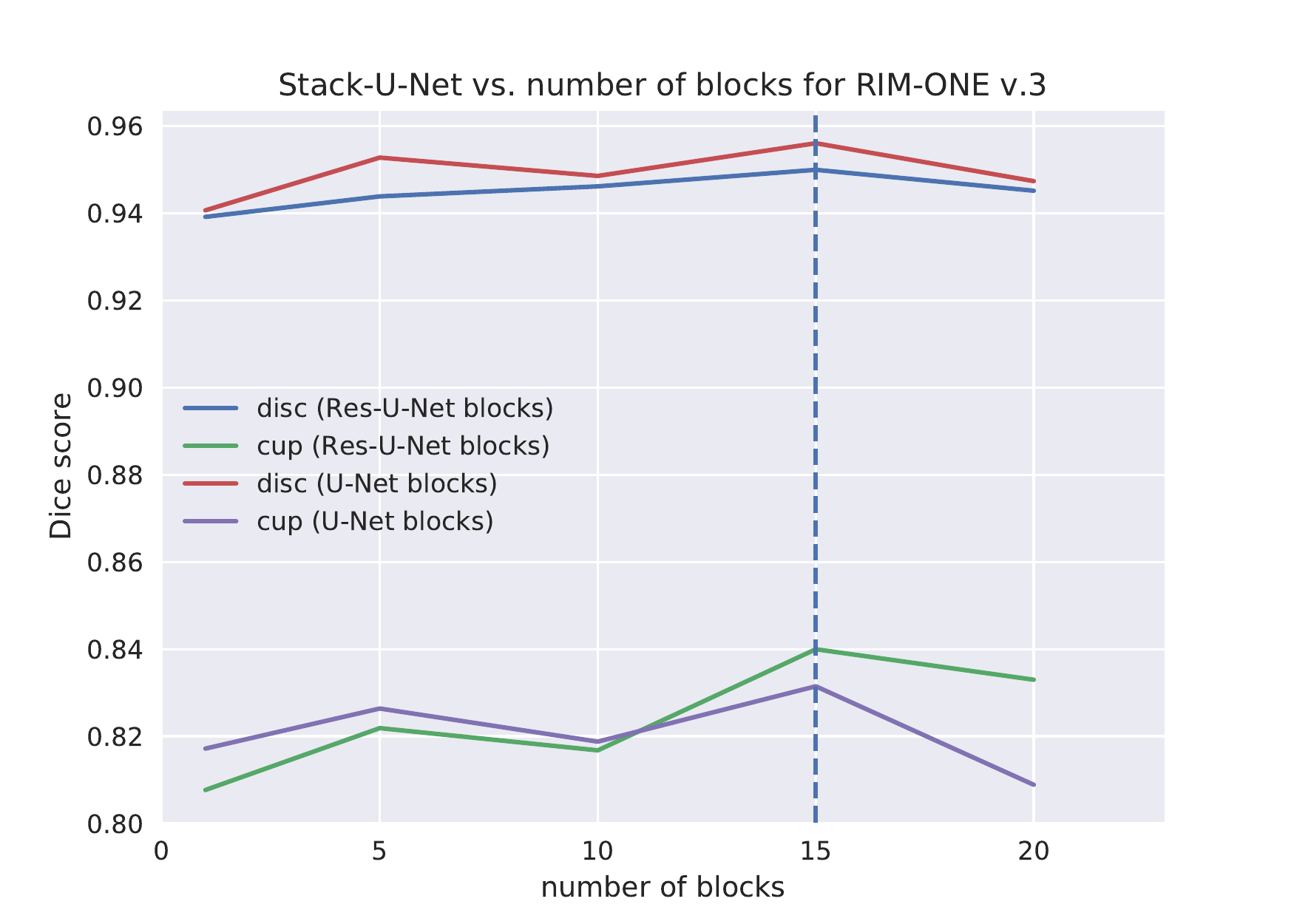}
        \caption{Stack-U-Net performance w.r.t. the number of basic blocks.}
        \label{fig:stack-u-net_vs_n_blocks}
    \end{figure}
    
    \begin{table}[ht!]
        	\centering
        	\begin{tabular}{l|c|c|c|c}
        		\hline
        		&
        		\multicolumn{4}{c}{\,RIM-ONE v.3\,} \\   
        		& \multicolumn{2}{c|}{\,Disc\,}  & \multicolumn{2}{c}{\,Cup\,}  \\ 
                & \,IOU\, & \,Dice\,  & \,IOU\, & \,Dice\, \\ \hline \hline
        		Stack-U-Net (15 Res-U-Net blocks) w/ skip & \textbf{0.91} & \textbf{0.95}& \textbf{0.73} & \textbf{0.84} \\
                Stack-U-Net (15 Res-U-Net blocks) w/o skip & 0.90 & 0.94 & 0.72 & 0.83 \\ \hline
                Stack-U-Net (15 U-Net blocks) w/ skip &\textbf{0.92} & \textbf{0.96} & 0.72 & 0.83 \\
                Stack-U-Net (15 U-Net blocks) w/o skip & 0.91 & 0.95 & \textbf{0.74} & \textbf{0.85} \\ \hline
        	\end{tabular}
            \newline\newline
        	\caption{Comparison of the cascade model with and without long skip connections linking input image with the first layer of each basic block.}
            \label{table:skip_connections}
\end{table}
        
\begin{table}[ht!]
            \centering
        	\begin{tabular}{l|c|c|c|c}
        		\hline
                & \multicolumn{4}{c}{\,UCSF-DB\,} \\   
                & \multicolumn{2}{c}{\,Disc\,}  & \multicolumn{2}{|c}{\,Cup\,}  \\ 
                & \,IOU\, & \,Dice\,  & \,IOU\, & \,Dice\, \\ \hline \hline 
                Stack-U-Net (15 Res-U-Net blocks) &  \textbf{0.92} & \textbf{0.96} & 0.73  & 0.84  \\
                Stack-U-Net (15 U-Net blocks) & \textbf{0.92} & \textbf{0.96} & \textbf{0.74}  & \textbf{0.85} \\
                U-Net &  \textbf{0.92} & 0.94  &  0.73  & 0.84 \\ 
                Mean Human-vs.-Human & 0.81 & 0.87 & 0.53 & 0.66 \\ \hline
            \end{tabular}
            \newline\newline
            \caption{Results on UCSF-DB large private dataset.}
            \label{table:ucsf_results}
\end{table}

We observe that the model with 15 blocks works better than with the lower and higher number of blocks, regardless of the block type (Fig.~\ref{fig:stack-u-net_vs_n_blocks}). Skip connections typically enhance the results by a small extent, except for the case of Stack-U-Net with 15 U-Net blocks without skip connections (Table~\ref{table:skip_connections}).

\begin{figure}[h!]
	\begin{subfigure}[t]{0.5\textwidth}
		\begin{subfigure}[t]{0.3\textwidth}
			\includegraphics[width=1.0\linewidth]{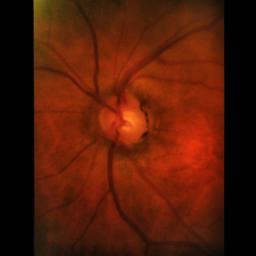} 
			\caption*{Input image}
		\end{subfigure}
		\begin{subfigure}[t]{0.3\textwidth}
			\includegraphics[width=1.0\linewidth]{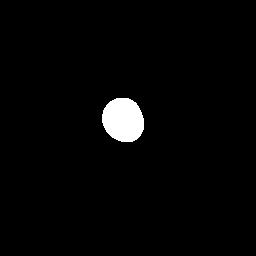} 
			\caption*{Predicted}
		\end{subfigure}
		\begin{subfigure}[t]{0.3\textwidth}
			\includegraphics[width=1.0\linewidth]{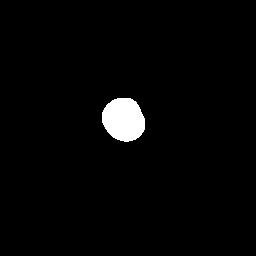} 
			\caption*{Correct}
		\end{subfigure}
		\caption*{\textbf{Disc}: best case (IOU = 0.96)}
	\end{subfigure}
	\hspace{0.15cm}
	\begin{subfigure}[t]{0.5\textwidth}
		\begin{subfigure}[t]{0.3\textwidth}
			\includegraphics[width=1.0\linewidth]{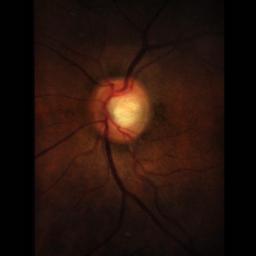} 
			\caption*{Input image}
		\end{subfigure}
		\begin{subfigure}[t]{0.3\textwidth}
			\includegraphics[width=1.0\linewidth]{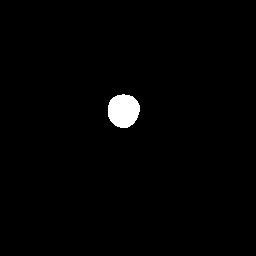} 
			\caption*{Predicted}
		\end{subfigure}
		\begin{subfigure}[t]{0.3\textwidth}
			\includegraphics[width=1.0\linewidth]{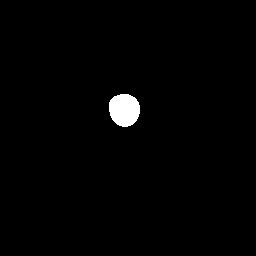} 
			\caption*{Correct}
		\end{subfigure}
		\caption*{\textbf{Cup}: best case (IOU = 0.91)}
	\end{subfigure}
	\vspace{0.2cm}
	
	\begin{subfigure}[t]{0.5\textwidth}
		\begin{subfigure}[t]{0.3\textwidth}
			\includegraphics[width=1.0\linewidth]{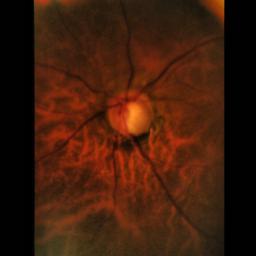} 
			\caption*{Input image}
		\end{subfigure}
		\begin{subfigure}[t]{0.3\textwidth}
			\includegraphics[width=1.0\linewidth]{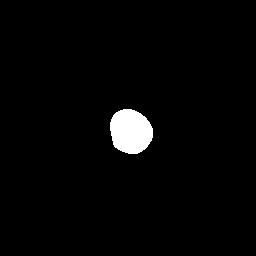} 
			\caption*{Predicted}
		\end{subfigure}
		\begin{subfigure}[t]{0.3\textwidth}
			\includegraphics[width=1.0\linewidth]{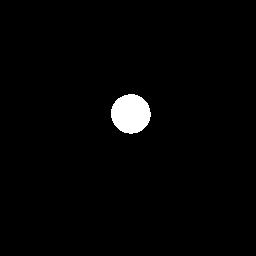} 
			\caption*{Correct}
		\end{subfigure}
		\caption*{\textbf{Disc}: worst case (IOU = 0.80)}
	\end{subfigure}
	\hspace{0.15cm}
	\begin{subfigure}[t]{0.5\textwidth}
		\begin{subfigure}[t]{0.3\textwidth}
			\includegraphics[width=1.0\linewidth]{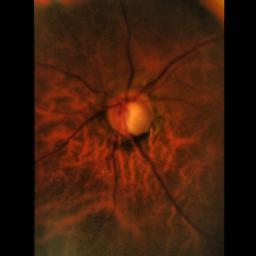} 
			\caption*{Input image}
		\end{subfigure}
		\begin{subfigure}[t]{0.3\textwidth}
			\includegraphics[width=1.0\linewidth]{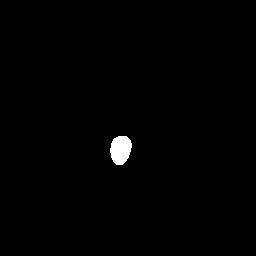} 
			\caption*{Predicted}
		\end{subfigure}
		\begin{subfigure}[t]{0.3\textwidth}
			\includegraphics[width=1.0\linewidth]{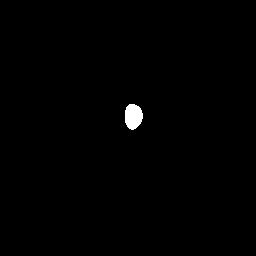} 
			\caption*{Correct}
		\end{subfigure}
		\caption*{\textbf{Cup}: worst case (IOU = 0.45)}
	\end{subfigure}
	
	\caption{The best and the worst cases of the algorithm performance on RIM-ONE v.3 database for the respective best models: \\ for optic disc --- with Stack-U-Net with 15 U-Net blocks, \\ for optic cup --- with Stack-U-Net with 15 Res-U-Net blocks.}
	\label{fig:visual_results}
\end{figure}

Visual comparison of the best and worst cases for the best-performing networks on each task for RIM-ONE v.3 database can be made based on Fig.~\ref{fig:visual_results}.
 

\section{Discussion}

We present the model for image segmentation based on a stack of the well-known U-Net models. Each model in a cascade refines the result of the previous one, directly accessing the colors from an input image. For the task of optic disc and optic cup segmentation on eye fundus image, which requires a solution for the reliable glaucoma detection, we report high results, and the model outperforms existing solutions by a large number of benchmarks.

Linear increase of the number of parameters and of the time of the forward / backward pass remains a drawback, and, together with the observed quality gap, it especially motivates the further research. 


\section*{Acknowledgment}


Blake M. Snyder was supported in part by the Doris Duke Charitable Foundation through a grant supporting the Doris Duke International Clinical Research Fellows Program at the University of California San Francisco School of Medicine.  Blake M. Snyder is a Doris Duke International Clinical Research Fellow. 

\bibliographystyle{splncs}
\bibliography{main}

\begin{thebibliography}{10}

\bibitem{med_almazroa}
Almazroa, A., Burman, R., Raahemifar, K., Lakshminarayanan, V.:
\newblock Optic disc and optic cup segmentation methodologies for glaucoma
  image detection: a survey.
\newblock Journal of ophthalmology \textbf{2015} (2015)

\bibitem{med_quigley}
Quigley, H.A., Broman, A.T.:
\newblock The number of people with glaucoma worldwide in 2010 and 2020.
\newblock British journal of ophthalmology \textbf{90}(3) (2006)  262--267

\bibitem{akram}
Akram, M.U., Tariq, A., Khalid, S., Javed, M.Y., Abbas, S., Yasin, U.U.:
\newblock Glaucoma detection using novel optic disc localization, hybrid
  feature set and classification techniques.
\newblock Australasian physical \& engineering sciences in medicine
  \textbf{38}(4) (2015)  643--655

\bibitem{lim2015integrated}
Lim, G., Cheng, Y., Hsu, W., Lee, M.L.:
\newblock Integrated optic disc and cup segmentation with deep learning.
\newblock In: Tools with Artificial Intelligence (ICTAI), 2015 IEEE 27th
  International Conference on, IEEE (2015)  162--169

\bibitem{milletari2016v}
Milletari, F., Navab, N., Ahmadi, S.A.:
\newblock V-net: Fully convolutional neural networks for volumetric medical
  image segmentation.
\newblock In: 3D Vision (3DV), 2016 Fourth International Conference on, IEEE
  (2016)  565--571

\bibitem{ronneberger2015u}
Ronneberger, O., Fischer, P., Brox, T.:
\newblock U-net: Convolutional networks for biomedical image segmentation.
\newblock In: International Conference on Medical image computing and
  computer-assisted intervention, Springer (2015)  234--241

\bibitem{lin2017refinenet}
Lin, G., Milan, A., Shen, C., Reid, I.:
\newblock Refinenet: Multi-path refinement networks for high-resolution
  semantic segmentation.
\newblock In: IEEE Conference on Computer Vision and Pattern Recognition
  (CVPR). (2017)

\bibitem{resnet}
He, K., Zhang, X., Ren, S., Sun, J.:
\newblock Deep residual learning for image recognition.
\newblock In: Proceedings of the IEEE conference on computer vision and pattern
  recognition. (2016)  770--778

\bibitem{imagenet}
Russakovsky, O., Deng, J., Su, H., Krause, J., Satheesh, S., Ma, S., Huang, Z.,
  Karpathy, A., Khosla, A., Bernstein, M.,  et~al.:
\newblock Imagenet large scale visual recognition challenge.
\newblock International Journal of Computer Vision \textbf{115}(3) (2015)
  211--252

\bibitem{dai2016instance}
Dai, J., He, K., Sun, J.:
\newblock Instance-aware semantic segmentation via multi-task network cascades.
\newblock In: Proceedings of the IEEE Conference on Computer Vision and Pattern
  Recognition. (2016)  3150--3158

\bibitem{christ2016automatic}
Christ, P.F., Elshaer, M.E.A., Ettlinger, F., Tatavarty, S., Bickel, M., Bilic,
  P., Rempfler, M., Armbruster, M., Hofmann, F., D’Anastasi, M.,  et~al.:
\newblock Automatic liver and lesion segmentation in ct using cascaded fully
  convolutional neural networks and 3d conditional random fields.
\newblock In: International Conference on Medical Image Computing and
  Computer-Assisted Intervention, Springer (2016)  415--423

\bibitem{sevastopolsky2017optic}
Sevastopolsky, A.:
\newblock Optic disc and cup segmentation methods for glaucoma detection with
  modification of u-net convolutional neural network.
\newblock Pattern Recognition and Image Analysis \textbf{27}(3) (2017)
  618--624

\bibitem{toshev2014deeppose}
Toshev, A., Szegedy, C.:
\newblock Deeppose: Human pose estimation via deep neural networks.
\newblock In: Proceedings of the IEEE conference on computer vision and pattern
  recognition. (2014)  1653--1660

\bibitem{trigeorgis2016mnemonic}
Trigeorgis, G., Snape, P., Nicolaou, M.A., Antonakos, E., Zafeiriou, S.:
\newblock Mnemonic descent method: A recurrent process applied for end-to-end
  face alignment.
\newblock In: Proceedings of the IEEE Conference on Computer Vision and Pattern
  Recognition. (2016)  4177--4187

\bibitem{szeliski2010computer}
Szeliski, R.:
\newblock Computer vision: algorithms and applications.
\newblock Springer Science \& Business Media (2010)

\bibitem{he2016identity}
He, K., Zhang, X., Ren, S., Sun, J.:
\newblock Identity mappings in deep residual networks.
\newblock In: European Conference on Computer Vision, Springer (2016)  630--645

\bibitem{drionsdb}
Carmona, E.J., Rinc{\'o}n, M., Garc{\'\i}a-Feijo{\'o}, J., Mart{\'\i}nez-de-la
  Casa, J.M.:
\newblock Identification of the optic nerve head with genetic algorithms.
\newblock Artificial Intelligence in Medicine \textbf{43}(3) (2008)  243--259

\bibitem{rimone}
Fumero, F., Alay{\'o}n, S., Sanchez, J., Sigut, J., Gonzalez-Hernandez, M.:
\newblock Rim-one: An open retinal image database for optic nerve evaluation.
\newblock In: Computer-Based Medical Systems (CBMS), 2011 24th International
  Symposium on, IEEE (2011)  1--6

\bibitem{drishtigs1}
Sivaswamy, J., Krishnadas, S., Chakravarty, A., Joshi, G., Tabish, A.S.,
  et~al.:
\newblock A comprehensive retinal image dataset for the assessment of glaucoma
  from the optic nerve head analysis.
\newblock JSM Biomedical Imaging Data Papers \textbf{2}(1) (2015)  1004

\bibitem{drishtigs2}
Sivaswamy, J., Krishnadas, S., Joshi, G.D., Jain, M., Tabish, A.U.S.:
\newblock Drishti-gs: Retinal image dataset for optic nerve head (onh)
  segmentation.
\newblock In: Biomedical Imaging (ISBI), 2014 IEEE 11th International Symposium
  on, IEEE (2014)  53--56

\bibitem{driu}
Maninis, K.K., Pont-Tuset, J., Arbel{\'a}ez, P., Van~Gool, L.:
\newblock Deep retinal image understanding.
\newblock In: International Conference on Medical Image Computing and
  Computer-Assisted Intervention, Springer (2016)  140--148

\bibitem{zilly2017glaucoma}
Zilly, J., Buhmann, J.M., Mahapatra, D.:
\newblock Glaucoma detection using entropy sampling and ensemble learning for
  automatic optic cup and disc segmentation.
\newblock Computerized Medical Imaging and Graphics \textbf{55} (2017)  28--41

\bibitem{zilly2015boosting}
Zilly, J.G., Buhmann, J.M., Mahapatra, D.:
\newblock Boosting convolutional filters with entropy sampling for optic cup
  and disc image segmentation from fundus images.
\newblock In: International Workshop on Machine Learning in Medical Imaging,
  Springer (2015)  136--143

\end{thebibliography}

\end{document}